\documentclass[final]{cvpr}
\usepackage{times}
\usepackage{epsfig}
\usepackage{graphicx}
\usepackage{amsmath}
\usepackage{amssymb}
\usepackage{cases}
\usepackage{diagbox}
\usepackage{algorithm}
\usepackage{algorithmic}
\usepackage{booktabs}
\usepackage{tablefootnote,footnotehyper}
\usepackage{multicol}
\usepackage{multirow}
\usepackage[accsupp]{axessibility}
\usepackage[pagebackref=false,breaklinks=true,letterpaper=true,colorlinks,urlcolor=blue,citecolor=green,linkcolor=blue,bookmarks=false]{hyperref}

\newcolumntype{L}[1]{>{\raggedright\arraybackslash}p{#1}}
\newcolumntype{C}[1]{>{\centering\arraybackslash}p{#1}}
\newcolumntype{R}[1]{>{\raggedleft\arraybackslash}p{#1}}

\def\cvprPaperID{5269} 
\def\confYear{CVPR 2022}

\title{Self-supervised Learning of Adversarial Example: \\Towards Good Generalizations for Deepfake Detection}

\begin{document}

\author{Liang Chen$^1$~~~~Yong Zhang$^{2}$\thanks{Corresponding authors. This work is done when L. Chen is an intern in Tencent AI Lab.}~~~~~~Yibing Song$^2$~~~~ Lingqiao Liu$^{1\ast}$~~~~Jue Wang$^{2}$\\
 {$^1$~The University of Adelaide}~~~~ {$^2$~Tencent AI Lab}\\
 {\tt\small \{liangchen527, zhangyong201303, yibingsong.cv, arphid\}@gmail.com} \\ {\tt\small lingqiao.liu@adelaide.edu.au}
}

\maketitle
\thispagestyle{empty}

\begin{abstract}
Recent studies in deepfake detection have yielded promising results when the training and testing face forgeries are from the same dataset. However, the problem remains challenging when one tries to generalize the detector to forgeries created by unseen methods in the training dataset.
This work addresses the generalizable deepfake detection from a simple principle: a generalizable representation should be sensitive to diverse types of forgeries. Following this principle, we propose to enrich the ``diversity'' of forgeries by synthesizing augmented forgeries with a pool of forgery configurations and strengthen the ``sensitivity'' to the forgeries by enforcing the model to predict the forgery configurations. To effectively explore the large forgery augmentation space, we further propose to use the adversarial training strategy to dynamically synthesize the most challenging forgeries to the current model.
Through extensive experiments, we show that the proposed strategies are surprisingly effective (see Figure \ref{fig auc}), and they could achieve superior performance than the current state-of-the-art methods. Code is available at \url{https://github.com/liangchen527/SLADD}.
\end{abstract}

\section{Introduction}

\begin{figure}[t]
    \centering
    \begin{tabular}{c}
          \includegraphics[width=0.98\linewidth, height=4.55 cm]{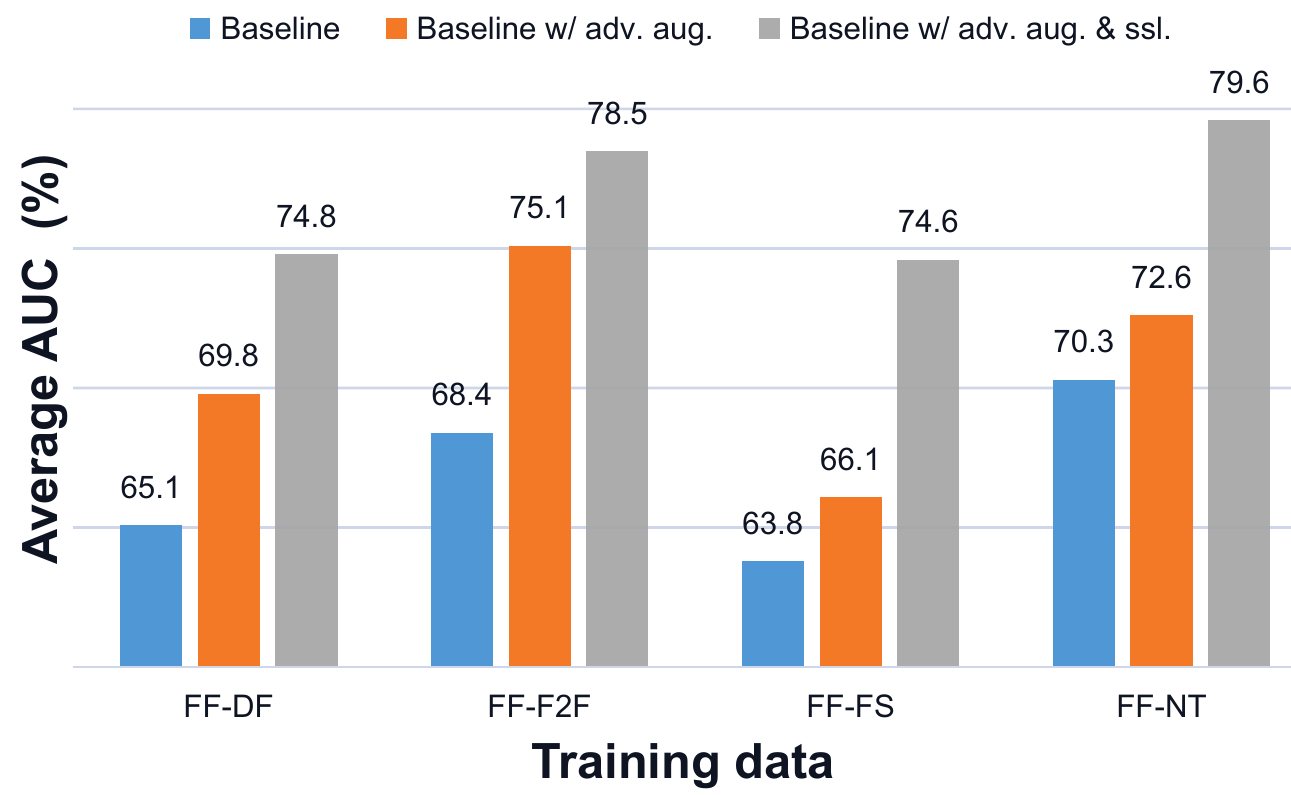} 
    \end{tabular}
    \caption{Performance improvements of the proposed strategies for the baseline model (\ie Xception \cite{rossler2019faceforensics++}). The models are trained on the four types of data from Faceforensics++ dataset \cite{rossler2019faceforensics++} and tested on CelebDF \cite{li2019celeb}, DFDC \cite{dfdc}, and DF1.0 \cite{jiang2020deeperforensics} datasets.}
	\label{fig auc}
	\vspace{-0.5 cm}
\end{figure}

The realistic image generation brought by the generative adversarial network (GAN) raises a security issue that human portraits can be easily substituted to provide malicious bioinformatics \cite{df,thies2016face2face,fs,thies2019deferred,wang2021high,zhang2020controllable,quan2022image,
yin2022styleheat,wang2021towards}. This forgery becomes a threat to subject identifications which have been extensively utilized in digital payment, video surveillance, and social media. To reduce these risks, there is an emerging investigation on deepfake detectors to identify face forgeries. Formulated as a binary classification problem, current detectors~\cite{li2018ictu, amerini2019deepfake, agarwal2019protecting, masi2020two, qian2020thinking} perform well when training and testing forgeries are synthesized from the same dataset and same forgery methods. However, in practice
the testing forgeries are usually from unseen datasets and synthesized by unseen methods. Discrepancies between training and testing data lead to inferior performance of detectors. There is a performance drop when detectors recognize forgeries outside the training dataset, which brings challenges to deepfake detectors for practical usage. 

Attempts have been made in recent arts to improve the generalizations.
For example, to overcome dataset bias, some studies \cite{li2020face, zhao2021learning} suggest data augmentation is an effective tool against poor generalization. These methods augment training data by synthesizing new face forgeries with their empirically designed augmentations. 
However, their augmentations are with a limited choice of strategy. The lacking of variety may jeopardize the generalization.
Meanwhile, intrinsic forgery attributes shared between forgeries, such as detail discrepancies~\cite{luo2021generalizing}, and frequency features~\cite{qian2020thinking,liu2021spatial}, are also mined broadly with hand-crafted representations to distinct forgeries.
But these attributes mainly rely on imperceptible image patterns, which are sensitive to post-processing steps, such as compression. They vary significantly in different datasets, thus leading to large detection bias and limiting their generalization \cite{zhu2021face}.

In this paper, we address the deepfake detection from a simple heuristic principle: \textit{a generalizable representation should be sensitive to the various types of forgeries.} Training a model by following this principle could potentially avoid the ``blind spot'' of the model or avoid relying on patterns specific to a dataset, since otherwise the model may not be capable of identifying a variety of forgeries. 
Pushing this idea to the limit, we propose to enrich the ``diversity of forgeries'' by synthesizing forgery images from a large pool of configurations \footnote{The term \textbf{configuration} means a specific way of synthesizing a forgery image. In the context of our discussion, it can also refer to a set of parameters that control a particular synthesizing process.}. Specifically, given a pristine, we randomly choose a reference image \footnote{We only use the forgery image in the training set as reference.} from training data, our synthesizer network (\ie generator) produces forgery configurations that specify the forgery region, the blending type, and the blending ratio. 
Then based on these configurations, a synthesized forgery is generated (some examples are shown in Figure~\ref{fig:teaser}). To enhance the ``sensitivity to the forgeries'', our detector network (\ie discriminator) is required to predict the configurations of an input in addition to judging if it is a forgery or not. Further, to effectively explore the large space of forgery augmentation, we adopt an adversarial training strategy to dynamically construct the augment that is most challenging to the current detector network. 
Different from the empirically designed augmentations \cite{li2020face, zhao2021learning, chen2022comen}, our adversarial augmentation strategy enjoys a large variety, and it can be dynamically constructed by the performance of the discriminator.

Through our experimental studies, we demonstrate the effectiveness of employing both adversarial augmentation and self-supervised tasks. A significant improvement over the baseline approach is observed, and our method also performs favorably against other state-of-the-art detectors. 

\section{Related Works}
\noindent{\flushleft \bf Deepfake detection.}
Recent studies have made various attempts for deepfake detection and achieve remarkable success \cite{li2018ictu, rossler2019faceforensics++, amerini2019deepfake, agarwal2019protecting, masi2020two, qian2020thinking, du2020towards, li2020face, luo2021generalizing, liu2021spatial, haliassos2021lips, wang2021representative, li2021frequency,nirkin2021deepfake,asnani2021reverse}. 
Several arts discuss low-level differences between pristines and forgeries and suggest using them as classification clues.
\def\swthree{0.3\linewidth}
\renewcommand{\tabcolsep}{1pt}
\begin{figure}[t]
\centering
    \begin{tabular}{ccc}
        \vspace{-0.5mm}\includegraphics[width=\swthree]{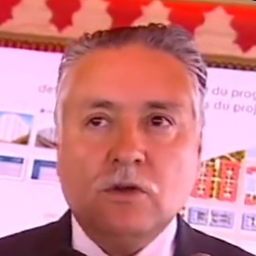}&\includegraphics[width=\swthree]{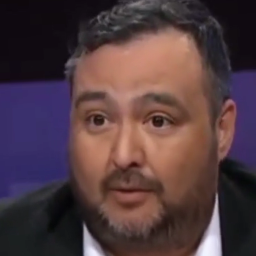}&
        \includegraphics[width=\swthree]{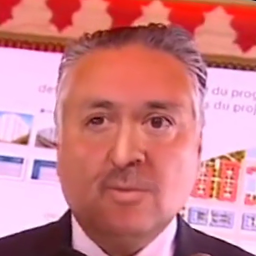}\\
        \vspace{-0.5mm}\includegraphics[width=\swthree]{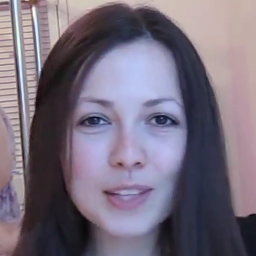}&
        \includegraphics[width=\swthree]{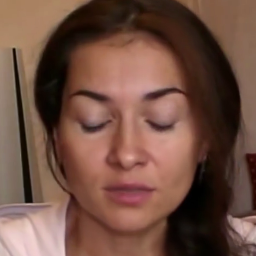}&
        \includegraphics[width=\swthree]{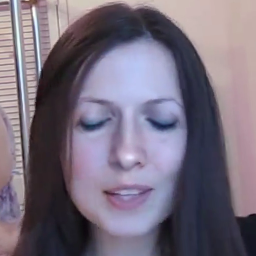}\\
        \includegraphics[width=\swthree]{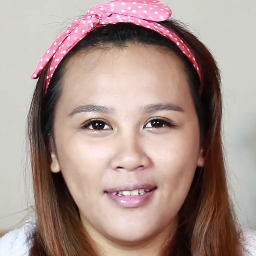}&
        \includegraphics[width=\swthree]{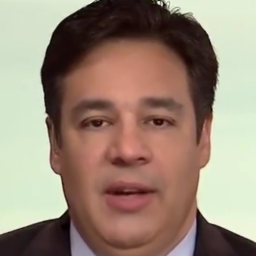}&
        \includegraphics[width=\swthree]{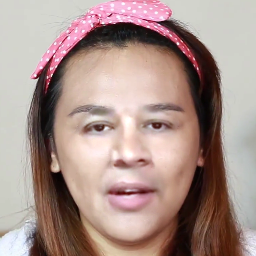}\\
        \small{\text{(a) Pristine}} & \small{\text{(b) Reference}}& \small{\text{(c) Forgery (adv)}}\\
    \end{tabular}
    \vspace{0.1 cm}
	\caption{Examples of the input pristine, reference images, and their corresponding synthesized adversarial forgeries.}
	\label{fig:teaser}
	\vspace{-0.5 cm}
\end{figure}
Li \etal \cite{li2020face} assume blending artifacts exist in all pristines and suggest finding the blending boundary beside the detection task;
Qian \etal \cite{qian2020thinking} and Luo \etal \cite{luo2021generalizing} use high-frequency details as additional inputs for their models;
Liu \etal \cite{liu2021spatial} adopt phase spectrums to capture the up-sampling artifacts of face forgery for the task.
Despite their effectiveness in many cases, the low-level artifacts are sensitive to post-processing steps which vary in different datasets, thus limiting their generalization \cite{zhu2021face}.
Some works suggest borrowing features from other tasks for deepfake detection.
Such as the features from lips reading \cite{haliassos2021lips}, facial image decomposition \cite{zhu2021face}, and landmark geometric features \cite{sun2021improving}.
Although these features can bring promising improvements, their generalization performance to the deepfake data are questionable. While annotating the deepfake data for these tasks is rather expensive, incorporating these tasks often lead to limited improvements.  

\begin{figure*}[t]
    \centering
    \begin{tabular}{c}
          \includegraphics[width=1\linewidth]{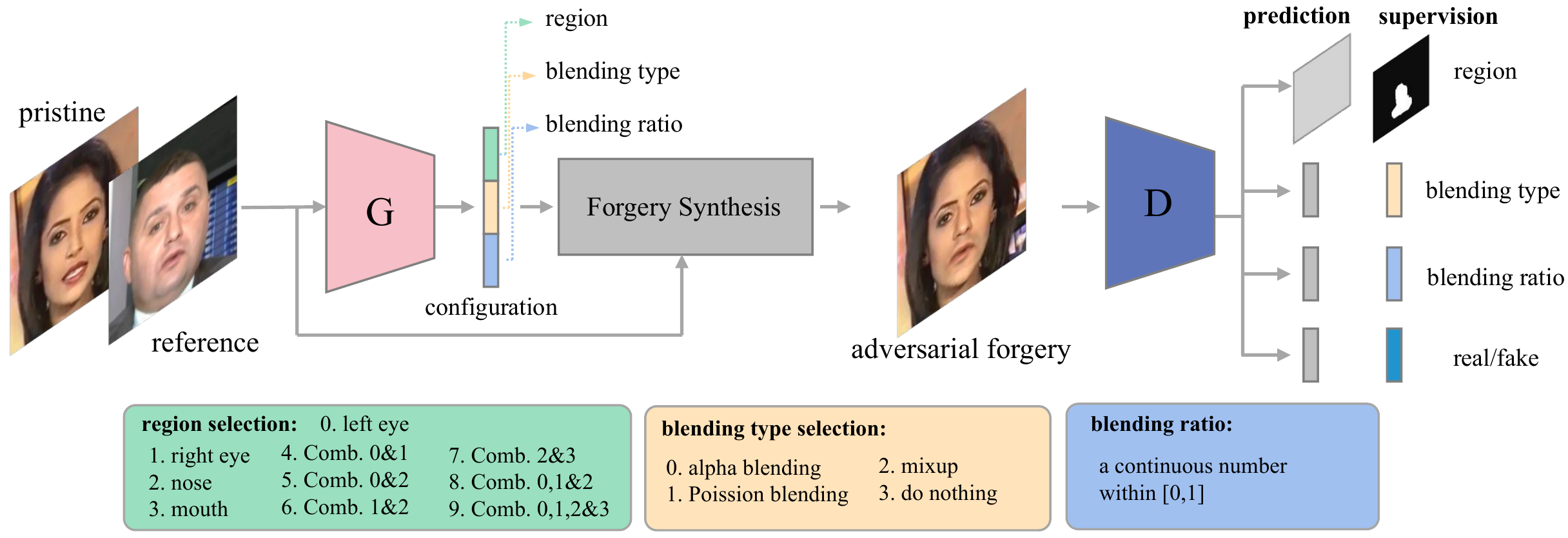} 
    \end{tabular}
    \caption{Overview of our model. The synthesizer network (\ie generator) outputs three forgery configurations that are further used to synthesize a new forgery, and these forgery configurations are also used as labels to guide the detector network (\ie discriminator). We train the generator and discriminator in an adversarial manner. Please refer to the text for details. 
    }
	\label{fig:pipeline}
	\vspace{-0.2 cm}
\end{figure*}
{\flushleft \bf Adversarial learning.}
Adversarial training aims to use adversarial examples for augmenting the training samples, which often contains a generator and a discriminator, and they are trained on the principle of defending against adversarial attacks \cite{goodfellow2014explaining, kurakin2016adversarial, madry2017towards, zhang2019adversarial}.
Goodfellow \etal \cite{goodfellow2014explaining} first use the fast gradient sign method to improve the adversarial robustness.
Madry \etal \cite{kurakin2016adversarial} further propose a multi-step scheme termed as projected gradient descent to enable powerful robustness compared to other works.
Moreover, different from previous works that use adversarial training to enlarge datasets through directly synthesizing new images \cite{tran2017bayesian,perez2017effectiveness,antoniou2017data,gurumurthy2017deligan,frid2018synthetic}, Zhang \etal \cite{zhang2019adversarial} suggest automatically choosing augmentation policies in an adversarial manner, which shows a great reduction in computing cost than the previous method \cite{cubuk2018autoaugment}.
However, most of these adversarial training strategies are designed for general tasks, such as classification.
In contrast, our adversarial example synthesizing process is similar to the deepfake generation procedure, thus is more suitable for the deepfake detection task.

\section{Proposed Method}
We propose to improve the generalizability of deepfake detector with the help of adversarial data augmentation, which enriches the types of forgeries, and self-supervised tasks that enforce sensitivity to forgery configurations.  
The pipeline of our model is shown in Figure \ref{fig:pipeline}. During training, we randomly select a reference image if the input is pristine. These two images are sent to a synthesizer network to produce configurations that specify the facial forgery region, the blending type, and the mixup blending ratio (if mixup blending is selected). Then a new forgery is synthesized based on these two images and the selected configurations. The synthesized forgery is passed to our detector for the real or fake and configurations predictions. 
Note the input will skip the forgery synthesizing process if it is an original forgery from the training data. 
We adopt an adversarial training strategy to train the system, in which the synthesizer is regarded as the generator, and the detector is regarded as the discriminator. The training process will train the forgery synthesizer to produce new forgeries to challenge the detector. Once trained, only the detector is used for deepfake detection.

\subsection{Selecting Space and Synthesizing Forgery}
\begin{figure*}
\scriptsize
\centering
	\begin{minipage}[b]{0.121\linewidth}
		\centering
		\centerline{
			\includegraphics[width =\linewidth]{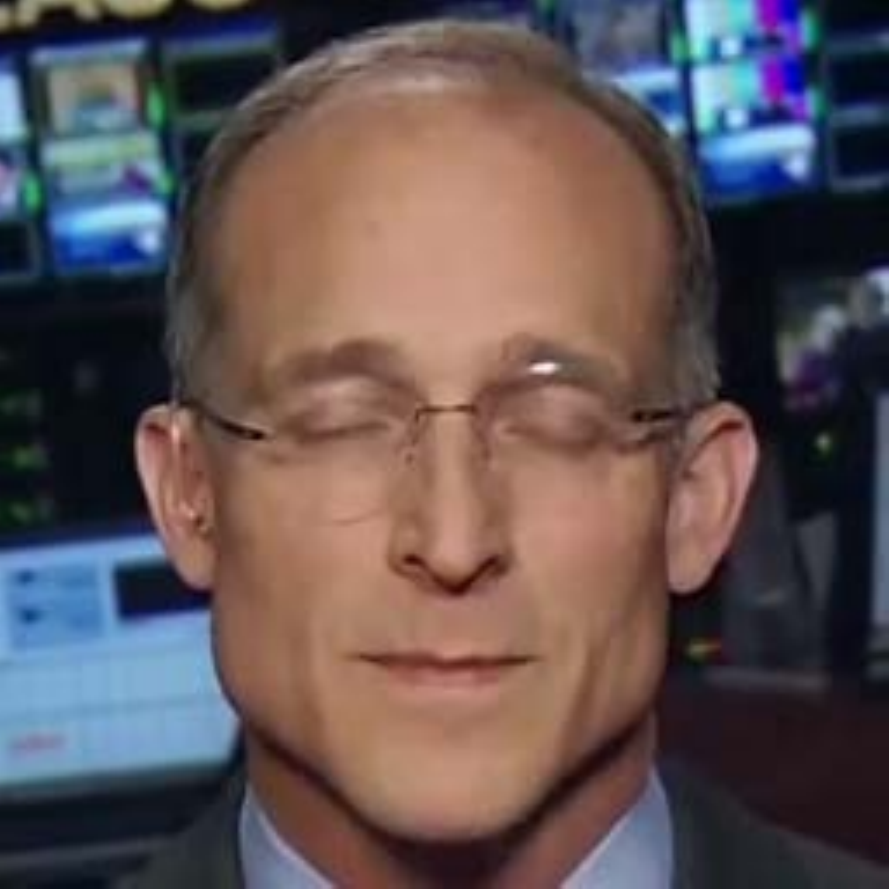}}
			\centerline{(a) Pristine}
			\centerline{}
	\end{minipage}
	\begin{minipage}[b]{0.121\linewidth}
		\centering
		\centerline{
			\includegraphics[width =\linewidth]{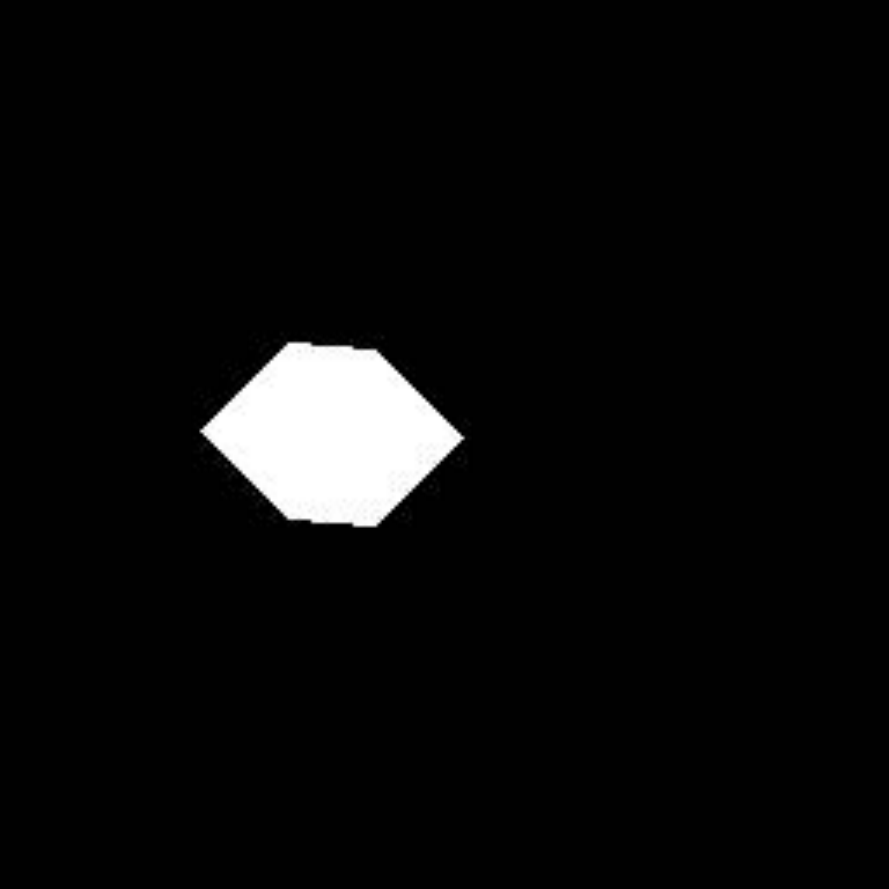}}
			\centerline{(c) No. 0.}
			\centerline{left eye}
	\end{minipage}
	\begin{minipage}[b]{0.121\linewidth}
		\centering
		\centerline{
			\includegraphics[width =\linewidth]{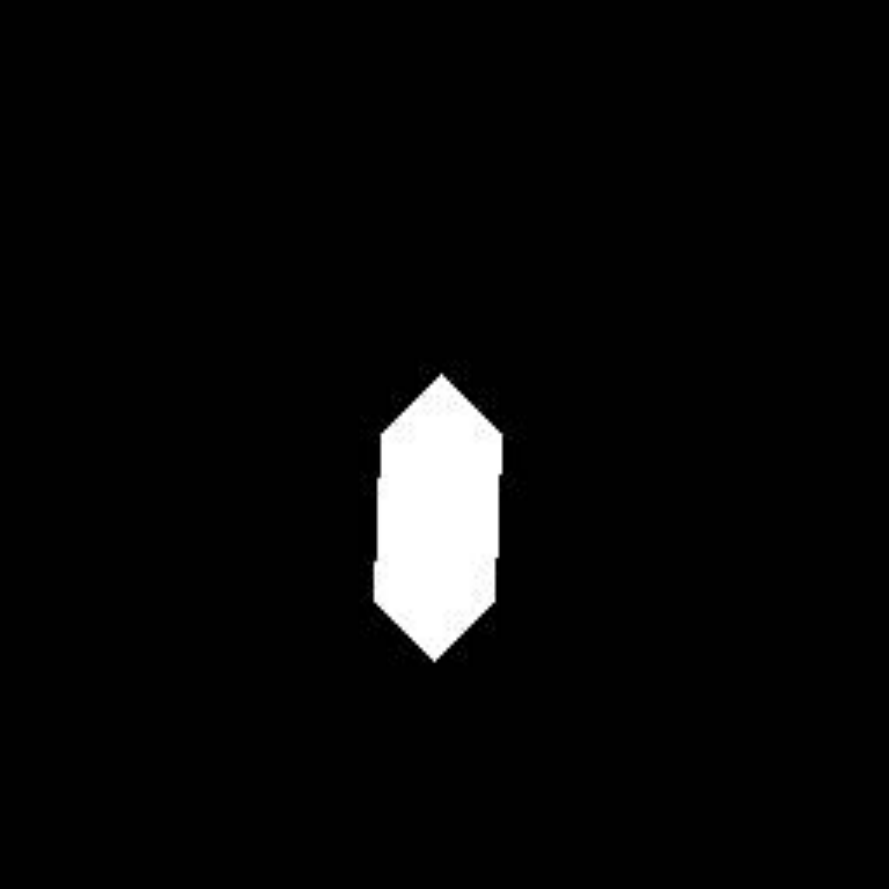}}
			\centerline{(e) No. 2.}
			\centerline{nose}
	\end{minipage}
	\begin{minipage}[b]{0.121\linewidth}
		\centering
		\centerline{
			\includegraphics[width =\linewidth]{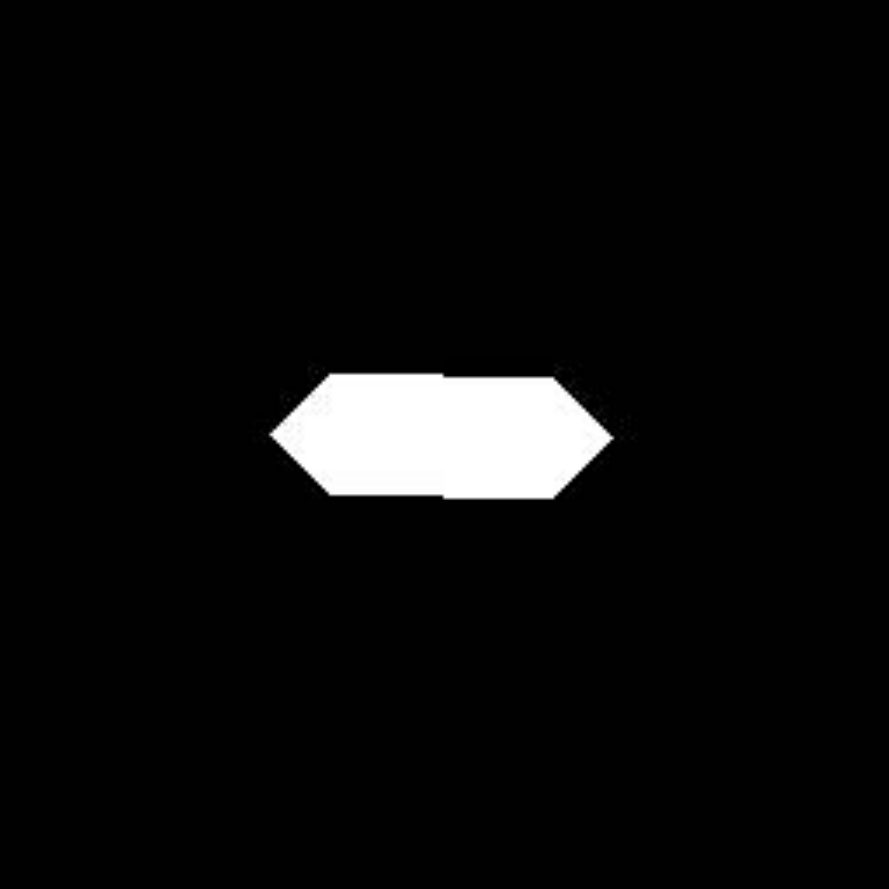}}
			\centerline{(g) No. 4.}
			\centerline{Comb. No. 0 \& 1}
	\end{minipage}
	\begin{minipage}[b]{0.121\linewidth}
		\centering
		\centerline{
			\includegraphics[width =\linewidth]{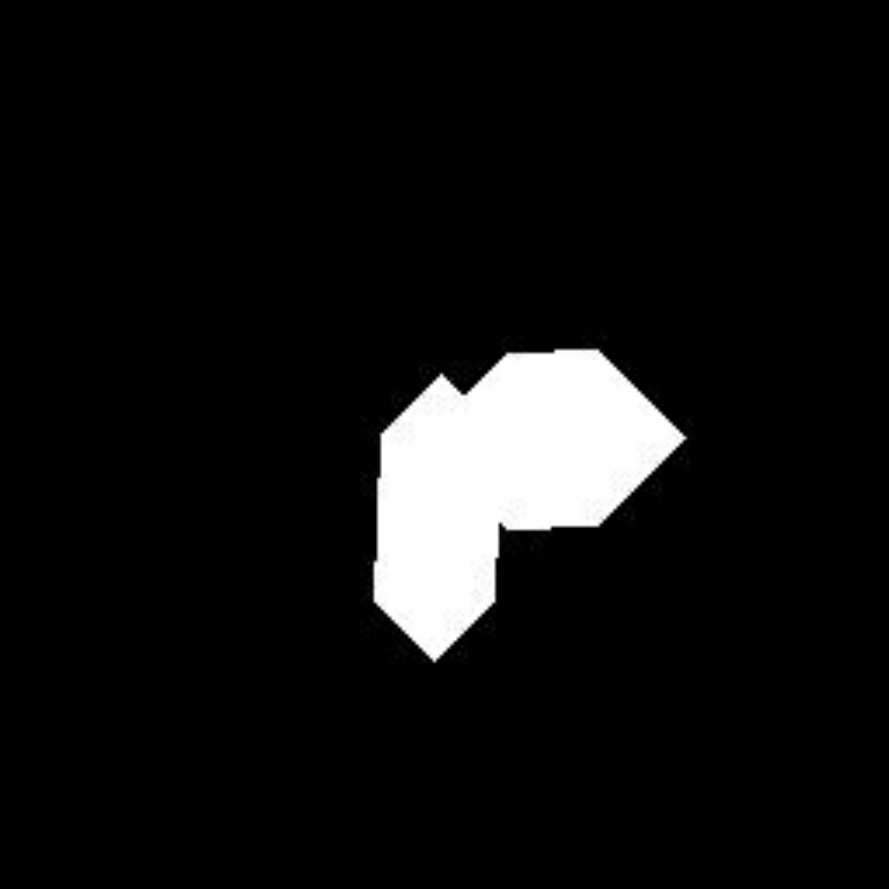}}
			\centerline{(i) No. 6.}
			\centerline{Comb. No. 1 \& 2}
	\end{minipage}
	\begin{minipage}[b]{0.121\linewidth}
		\centering
		\centerline{
			\includegraphics[width =\linewidth]{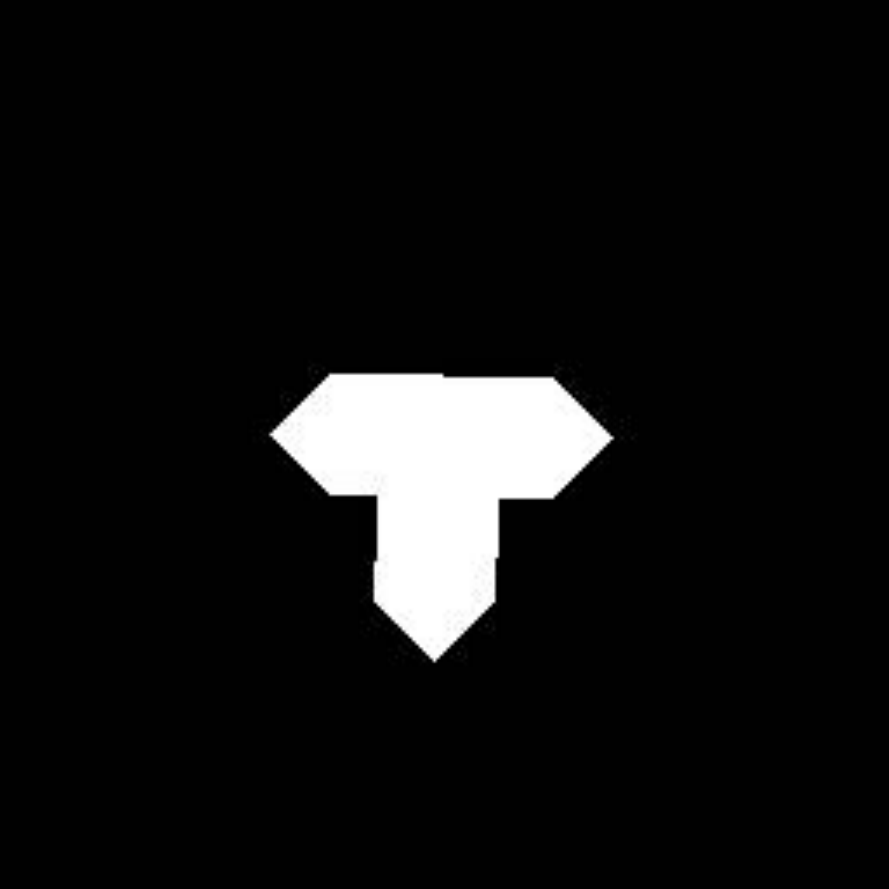}}
			\centerline{(k) No. 8.}
			\centerline{Comb. No. 0, 1 \& 2}
	\end{minipage}
	\begin{minipage}[b]{0.121\linewidth}
		\centering
		\centerline{
			\includegraphics[width =\linewidth]{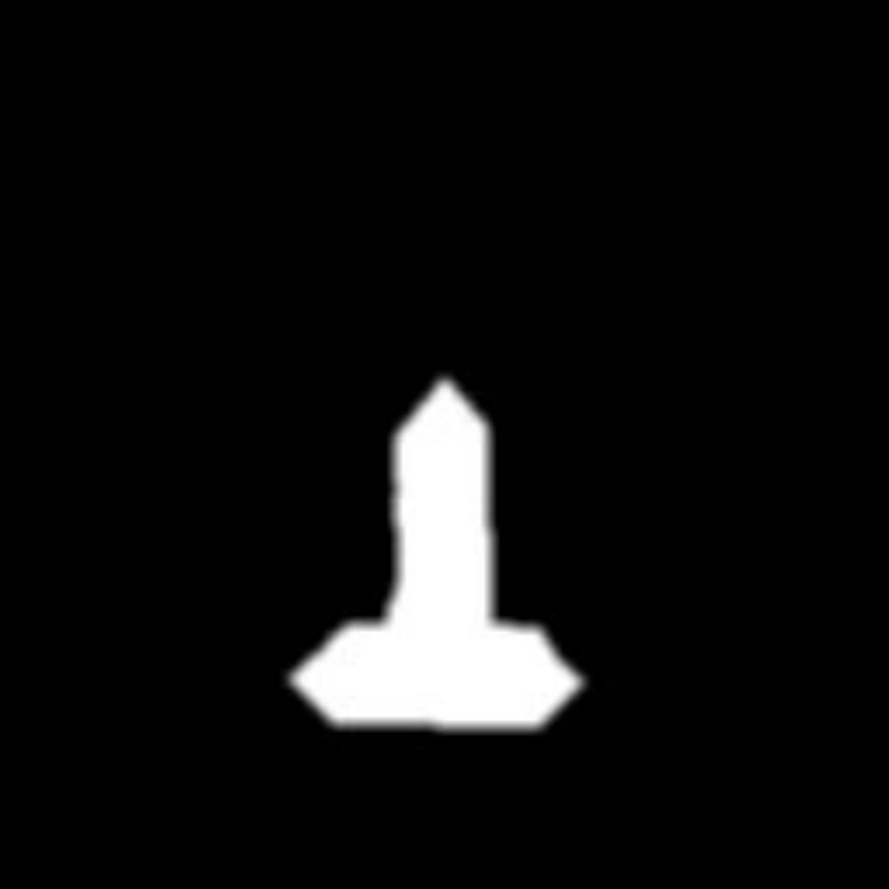}}
			\centerline{(m) deformed}
			\centerline{final mask}
	\end{minipage}
	\begin{minipage}[b]{0.121\linewidth}
		\centering
		\centerline{
			\includegraphics[width =\linewidth]{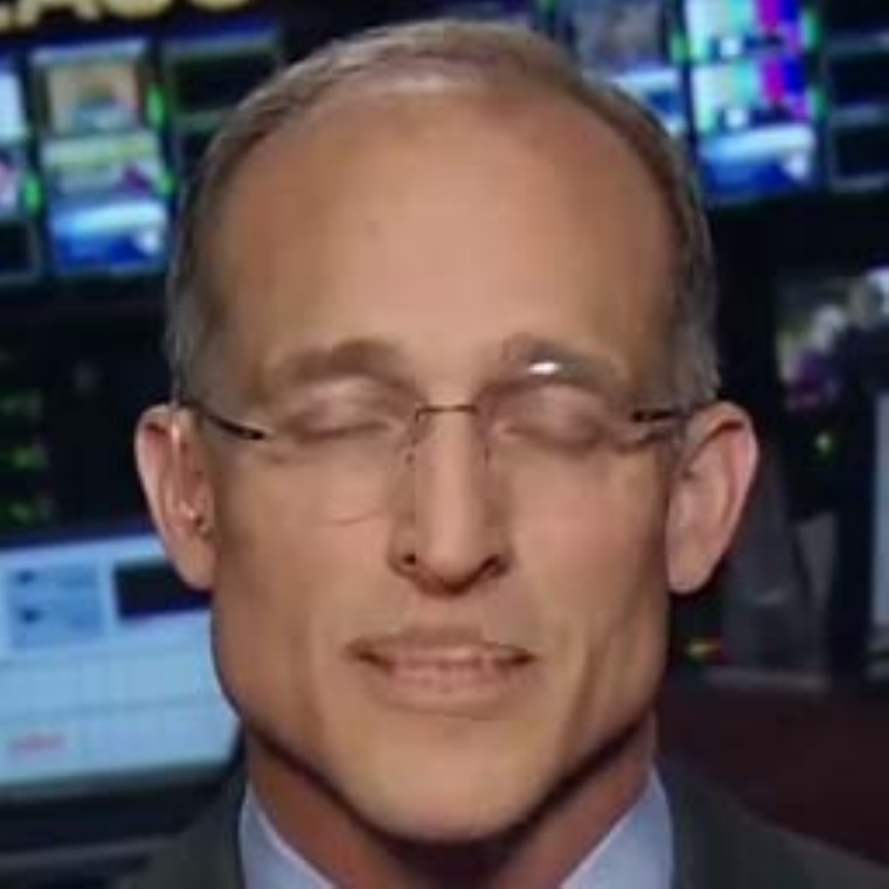}}
			\centerline{(o) Forgery by}
			\centerline{mixup blending}
	\end{minipage}\\
	\begin{minipage}[b]{0.121\linewidth}
		\centering
		\centerline{
			\includegraphics[width =\linewidth]{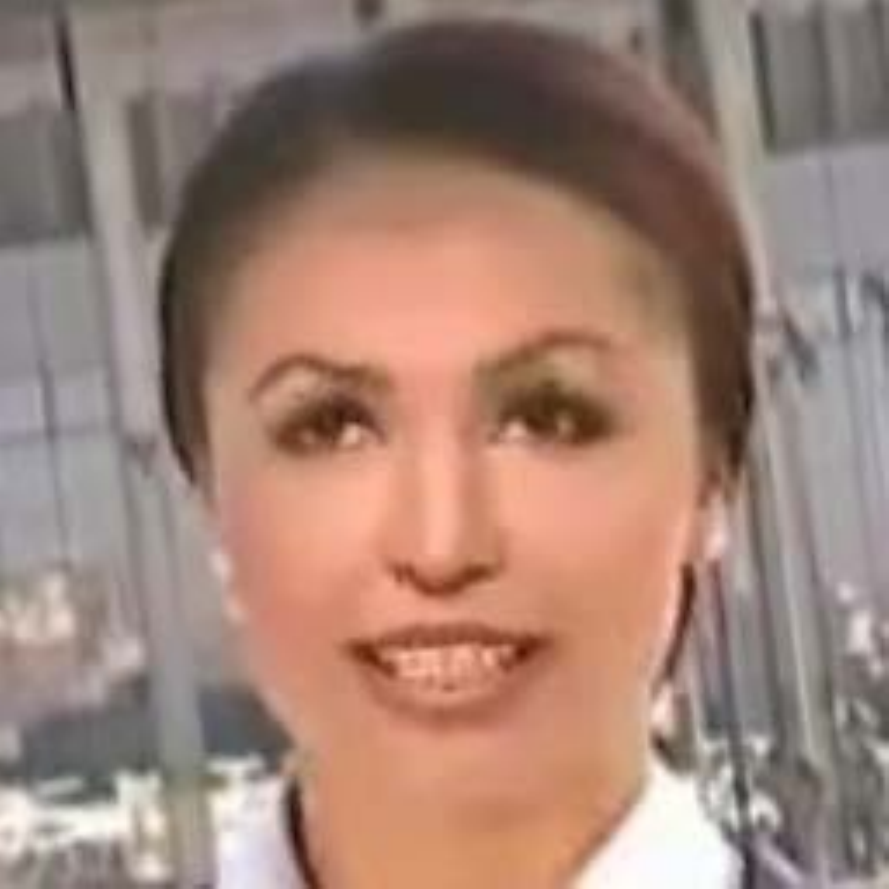}}
			\centerline{(b) Random reference}
			\centerline{}
	\end{minipage}
	\begin{minipage}[b]{0.121\linewidth}
		\centering
		\centerline{
			\includegraphics[width =\linewidth]{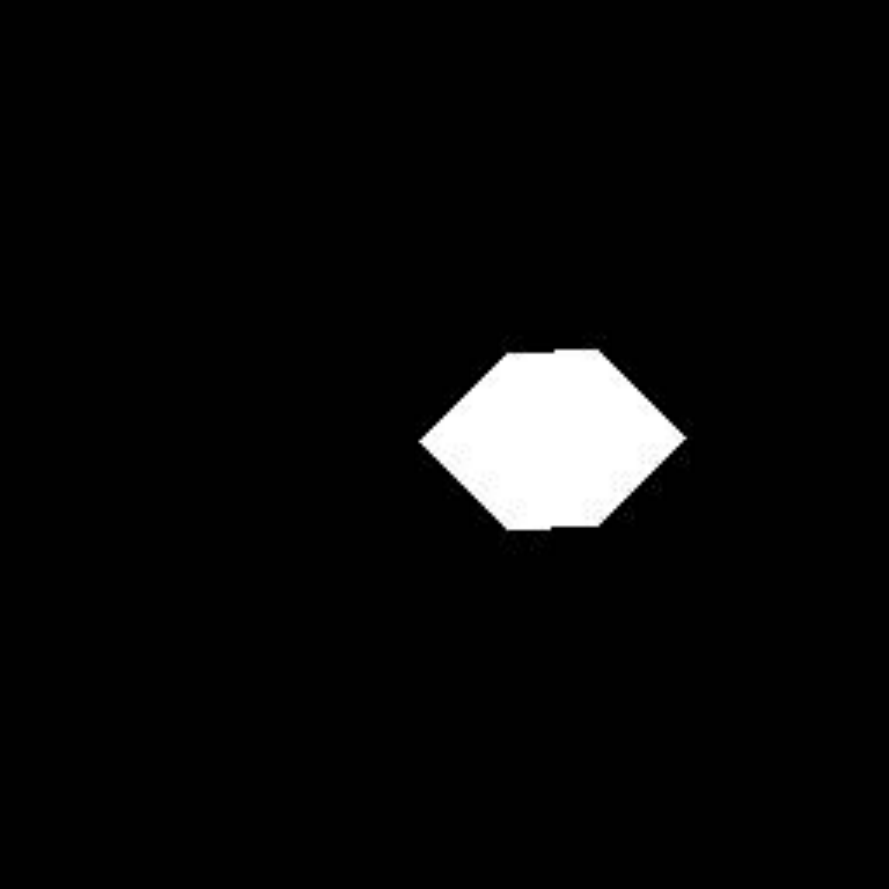}}
			\centerline{(d) No. 1.}
			\centerline{right eye}
	\end{minipage}
	\begin{minipage}[b]{0.121\linewidth}
		\centering
		\centerline{
			\includegraphics[width =\linewidth]{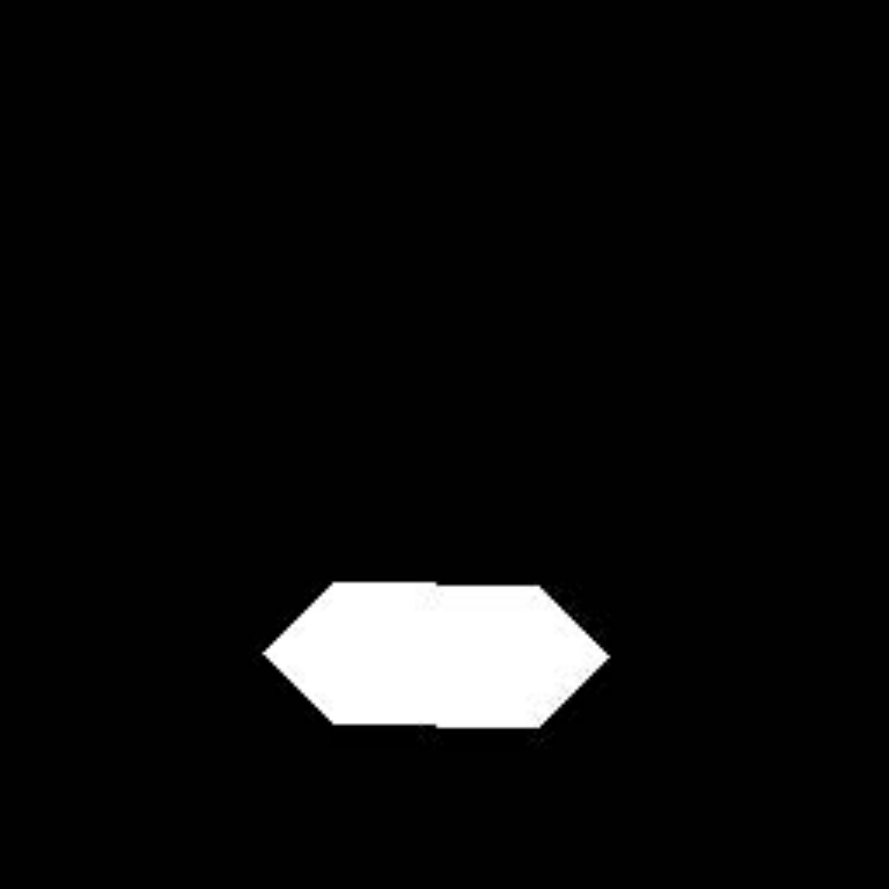}}
			\centerline{(f) No. 3.}
			\centerline{mouth}
	\end{minipage}
	\begin{minipage}[b]{0.121\linewidth}
		\centering
		\centerline{
			\includegraphics[width =\linewidth]{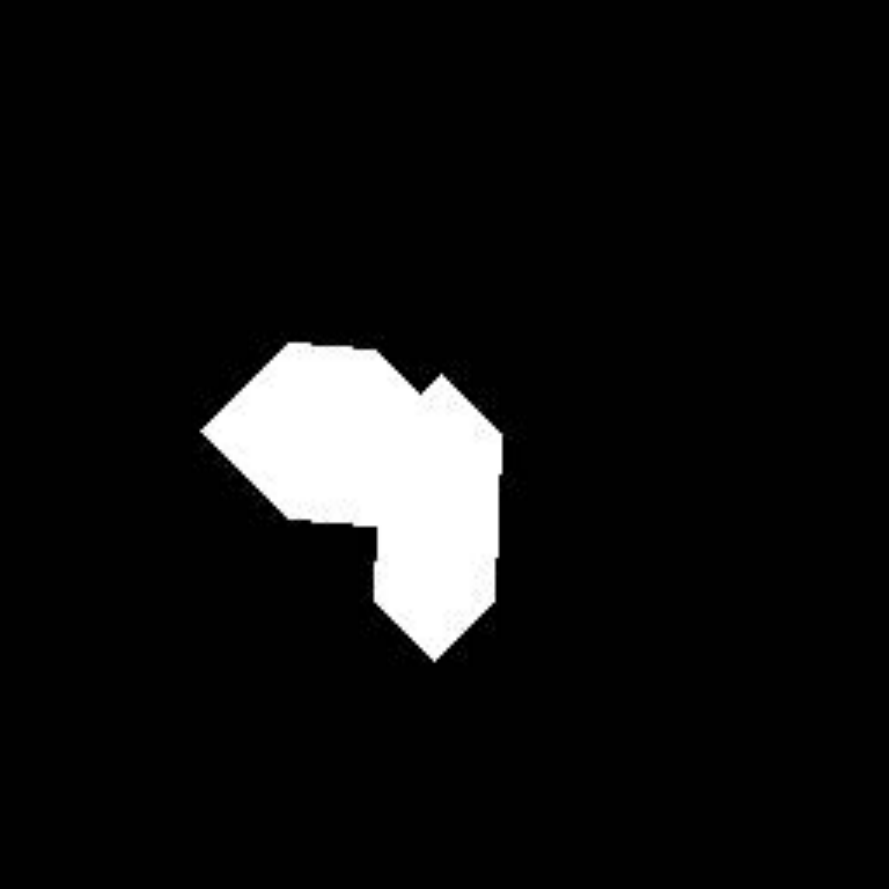}}
			\centerline{(h) No. 5.}
			\centerline{Comb. No. 0 \& 2}
	\end{minipage}
	\begin{minipage}[b]{0.121\linewidth}
		\centering
		\centerline{
			\includegraphics[width =\linewidth]{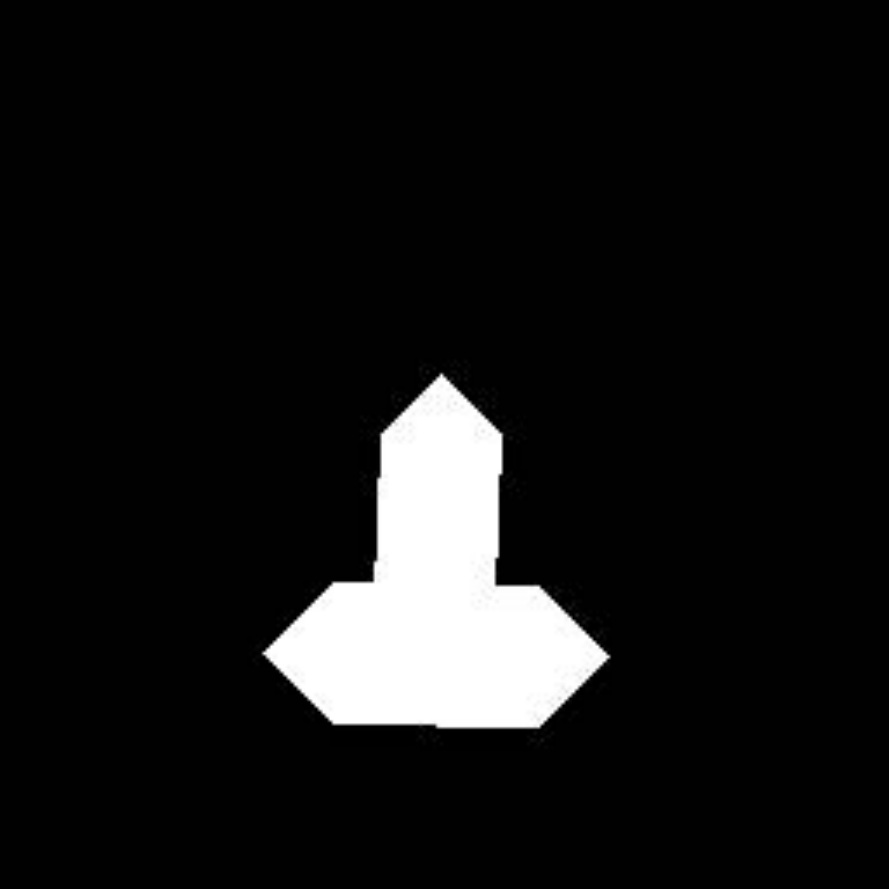}}
			\centerline{(j) No. 7.}
			\centerline{Comb. No. 2 \& 3}
	\end{minipage}
	\begin{minipage}[b]{0.121\linewidth}
		\centering
		\centerline{
			\includegraphics[width =\linewidth]{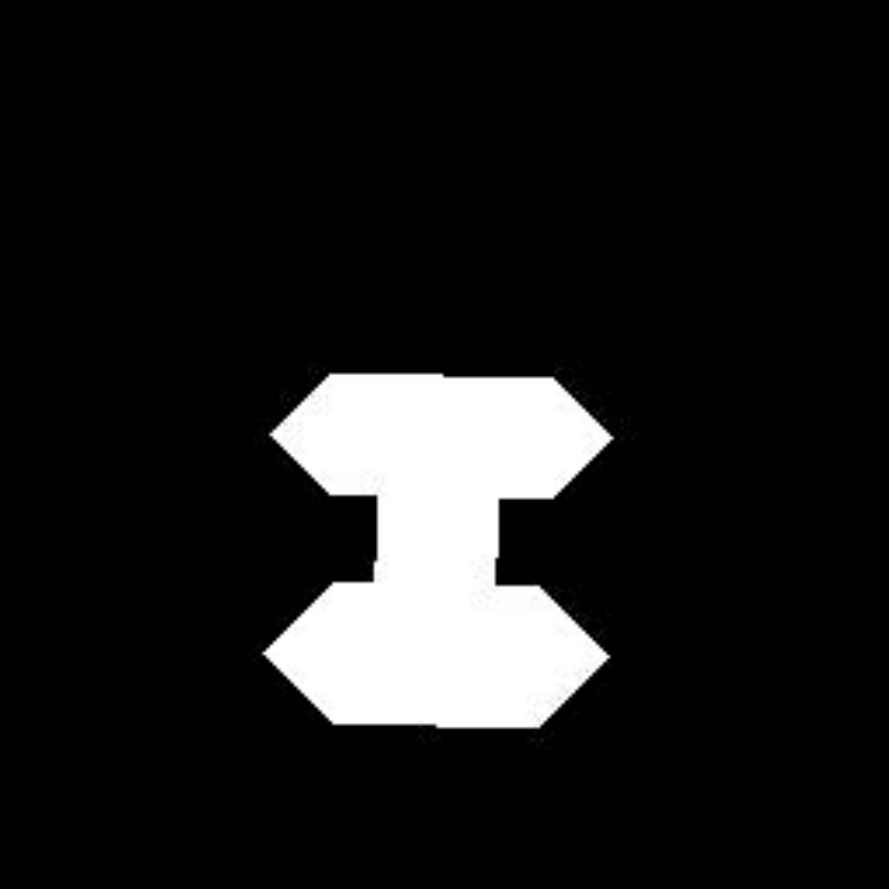}}
			\centerline{(l) No. 9.}
			\centerline{Comb. No. 0, 1, 2 \& 3}
	\end{minipage}
	\begin{minipage}[b]{0.121\linewidth}
		\centering
		\centerline{
			\includegraphics[width =\linewidth]{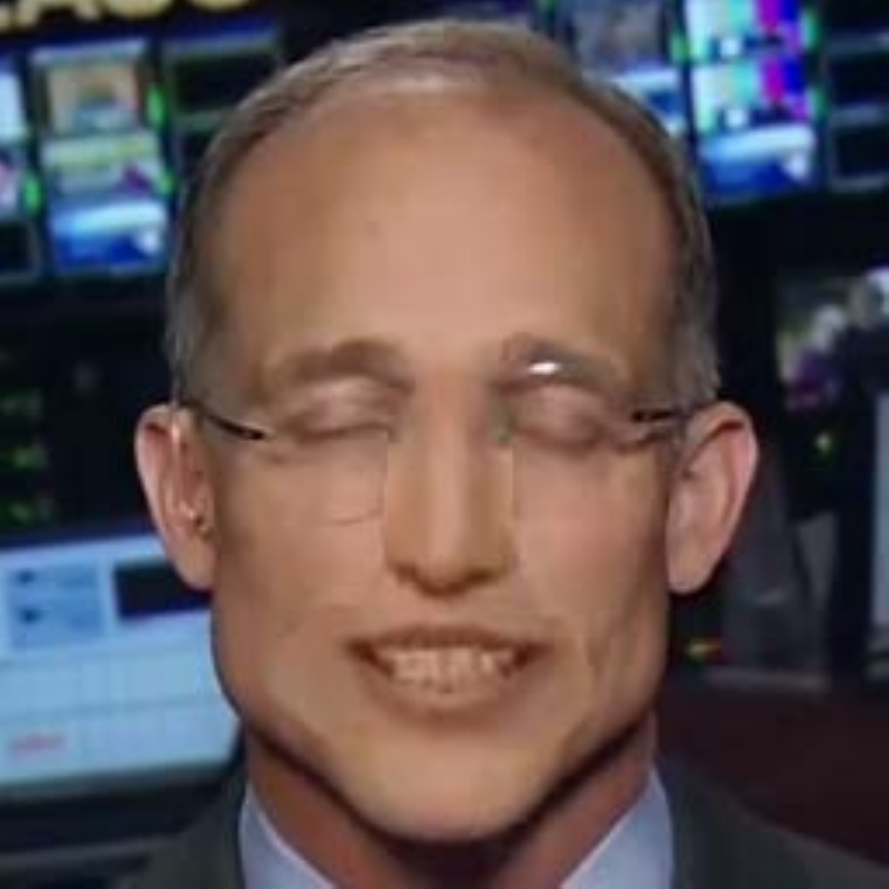}}
			\centerline{(n) Forgery by}
			\centerline{alpha blending}
	\end{minipage}
	\begin{minipage}[b]{0.121\linewidth}
		\centering
		\centerline{
			\includegraphics[width =\linewidth]{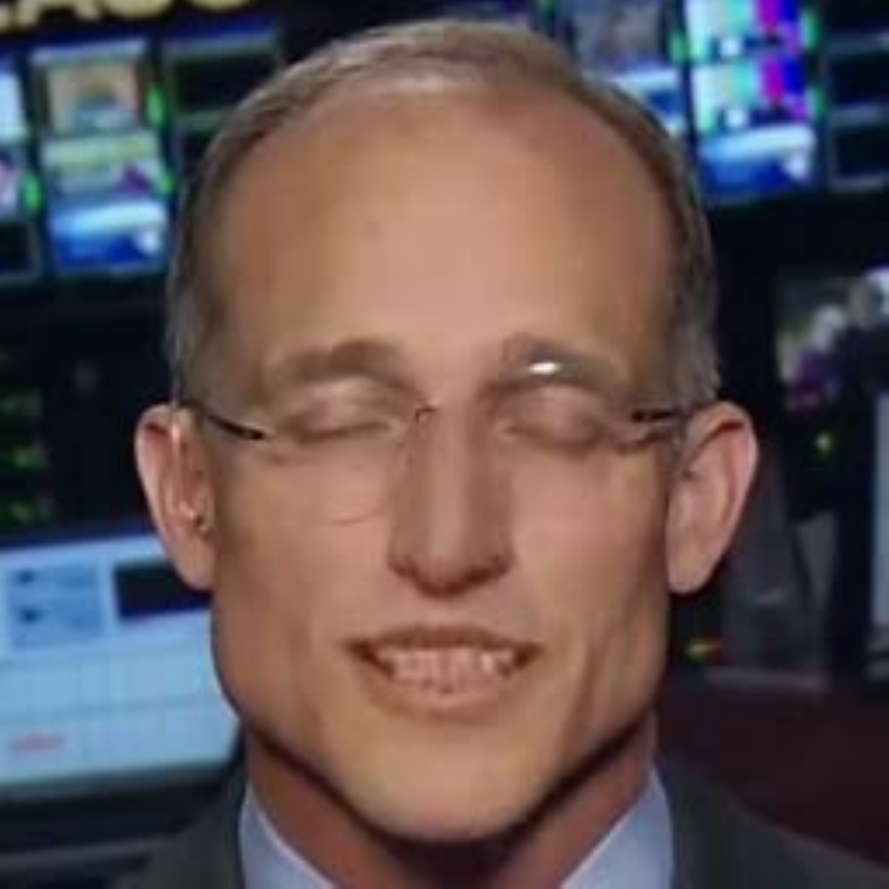}}
			\centerline{(p) Forgery by}
			\centerline{Poisson blending}
	\end{minipage}
	\caption{An example of the pristine and randomly chosen reference images (\ie (a) and (b)), the corresponding selecting space of the manipulate regions (\ie (c) - (m)), and the adversarial forgeries synthesized by different blending types (\ie (n) - (p)) under the instruction of the deformed manipulate region map (\ie (m)).
	The selected region number is 7. in this example, and the blending ratio for (o) is 0.5.}
	\label{fig controller}
	\vspace{-0.3 cm}
\end{figure*}

Our synthesizer network $G(\cdot, \theta)$ takes a pristine $\textbf{I}_p \in \mathbb{R}^{H \times W \times 3}$ and a reference image $\textbf{I}_f \in \mathbb{R}^{H \times W \times 3}$ as inputs and output three configurations: the forgery region reference index $R_g$, the blending type index $T_g$ (a discrete number), and the mix-up blending ratio $A_g$ (a continuous value). For $A_g$, we directly generate a scalar from the input $\textbf{I}_p$ and $\textbf{I}_f$. For $R_g$, $T_g$, we first generate two probability distributions $p(R_g)$ and $p(T_g)$, indicating the chance of selecting a particular index of $R_g$ and $T_g$, and then randomly sample a index according to the probability. For the simplicity of discussion, we use $p_m$ to denote those two probabilities.
These configurations, $R_g$, $T_g$ and $A_g$, control how a forgery image is generated. Their definitions are as follows:

\noindent{$R_g$:} The index $R_g \in \{0,\cdots,9\}$ determines a specific facial region. We divide the facial image into 10 regions based on its landmark, including the left eye, right eye, nose, mouth, and their combinations. Examples of different regions are shown in Figure \ref{fig controller} (c) - (l). 
We only consider the facial features because most deepfake arts focus on them, and they convey the most information in a facial image;

\noindent{$T_g$:} The reference number $T_g \in \{0,\cdots,3\}$ determines the blending type. We use three blending techniques, including alpha, Poisson \cite{perez2003poisson}, and mixup blending. Without the loss of generality, we include a do-nothing choice among the blending type pool, so that we can use the original pristine from the dataset for further classification.

\noindent{$A_g$:} The continuous value $A_g\in [0,1]$ is the blending ratio which is only effective when mixup blending is selected.

\setlength{\belowdisplayskip}{6pt} \setlength{\belowdisplayshortskip}{6pt}
\setlength{\abovedisplayskip}{6pt} \setlength{\abovedisplayshortskip}{6pt}

We adopt a random shape deformation operation and apply Gaussian blur with random kernel size to the selected forgery region before the synthesizing step. An example of the final mask is shown in Figure \ref{fig controller} (m).
Then, to synthesize the adversarial forgery $\textbf{I}_a$, we crop the facial parts determined by $R_g$ from $\textbf{I}_f$, and blend it into $\textbf{I}_p$ using the configurations indicated by the synthesizer network.
Specifically, we use the official implementations from OpenCV for alpha and Poisson blendings to synthesize $\textbf{I}_a$, and for mixup blending, $\textbf{I}_a$ can be obtained by:
\begin{equation}
\label{eq mixup_blending}
\textbf{I}_a = A_g \times \textbf{M}_{d}  \ast (\textbf{I}_f - \textbf{I}_p) + \textbf{I}_p,
\end{equation}
where $\textbf{M}_d$ is the deformed final mask, and $\ast$ is the Hadamard product.

Like existing deepfake methods, the cropped facial parts are applied with color transfer (aligning the mean of the RGB channels, respectively) and face alignment before the alpha and Poisson blending steps, which can avoid obvious artifacts in the newly synthesized forgery.
Examples of newly synthesized forgeries with different blending types are shown in Figure \ref{fig controller} (n) - (p).

\subsection{Joint Training with Self-Supervised Tasks}
\label{sec selfsupervise}
Self-supervised tasks have been shown effective for improving generalization in previous works \cite{chen2019self,wang2020learning}.
However, those attempts often consider self-supervised learning tasks proven effective for image classification tasks. In contrast to them, this paper employs an auxiliary self-supervised task specifically designed for the deepfake detection problem. By doing so, we expect that the auxiliary task aligns better with the target task and thus could lead to better performance.
Our idea follows a multi-task scheme that allows the model to simultaneously learn from two tasks, the main task and a set of auxiliary self-supervised task. Specifically, for the auxiliary tasks, we consider forgery region estimation, blending type estimation and blending ratio estimation. 
We elaborate on each task and the loss in the following.

{\flushleft \bf Main task loss $\mathcal{L}_{Main}$}.
The main task is a binary classification task that predicts whether a given input image is pristine or forgery.
Similar to previous work \cite{luo2021generalizing}, we adopt the AM-Softmax Loss \cite{wang2018additive} to compute $\mathcal{L}_{Main}$ since it allows for smaller intra-class variations and larger inter-class differences than the regular cross-entropy loss.
{\flushleft \bf Forgery region estimation loss $\mathcal{L}_{R}$}.
We also create a forgery region estimation task. Specifically, a forgery region prediction head is attached to the detector network and generates forgery region mask with the size of $[H/16 \times W/16]$. 
The ground-truth forgery mask $\textbf{M}_{gt}$ is created depending on categories of the input images, which is divided into three categories. 
Specifically, if the input image is the adversarial forgery, the corresponding $\textbf{M}_{gt}$ is the resized deformed final mask (\ie $\textbf{M}_{gt} = \textbf{M}_d$);
As most datasets provide the ground truth forgery region, we can directly use them as $\textbf{M}_{gt}$ if the input image is an original forgery from the training dataset;
If the input image is an original pristine from the training dataset, the $\textbf{M}_{gt}$ is an all-zero matrix (\ie $\textbf{M}_{gt} = \textbf{0}$), indicating there is no forgery region in the input image.
An $L_1$ loss is applied for the task:
\begin{equation}
\label{eq maploss}
\mathcal{L}_{R} = \frac{\Vert \textbf{M}_{gt} - \textbf{M}_e \Vert_1}{H/16 \times W/16},
\end{equation}
where $\textbf{M}_e$ is the estimated region map from our detector network, and $\Vert \cdot \Vert_1$ denotes the $L_1$ norm.
{\flushleft \bf Blending type estimation loss $\mathcal{L}_{T}$}.
We then introduce a blending type estimation task that aims to estimate the blending type of input images.
Similar to the forgery region estimation task, the ground-truth of the prediction target, i.e., ground-truth blending type $T_{gt}$, for this task varies accordingly to the category of the input data.
When the input is a synthesized adversarial forgery, $T_{gt}$ can be directly borrowed from the output of the generator (\ie $T_{gt} = T_g$, where $T_g$ $\in \{0,1,2\}$ according to the selected blending type); 
If the input is the original pristine, we set ${T}_{gt} = 3$, implying there is no blending operation in the image;
As the blending types are often unavailable in the existing datasets, we thus set $T_{gt} = 4$ if the input image is the original forgeries from the training dataset, indicating a blending type outside of our selecting space.
Similarly, we adopt the AM-Softmax Loss \cite{wang2018additive} to compute $\mathcal{L}_{T}$.
{\flushleft \bf Blending ratio estimation loss $\mathcal{L}_{A}$}.
To take full advantage of the blending information provided in the forgery synthesizing process, we further construct a task to estimate the blending ratio output by the synthesizer network.
Because the blending ratio is only effective when the synthesizer network selects the mixup blending, we can set the prediction ground-truth $A_{gt}={A}_g$ when the input image is a synthesized adversarial forgery using mixup blending. Correspondingly, this loss is not computed when the input image is either the original images in the training dataset or adversarial forgery synthesized with other blending types.
An $L_1$ loss is adopted to compute $\mathcal{L}_{A}$:
\begin{equation}
\mathcal{L}_{A} = \tau \times \Vert {A}_{gt}- A_e\Vert_1,  \label{eq blendingloss}
\end{equation}
where $A_e$ is the estimated blending ratio from the detector network, and $\tau$ is a binary value, which s.t. $\tau=1$ if $T_{gt}=2$ (\ie the blending type is mixup blending) and $\tau=0$ otherwise, thus ensuring $\mathcal{L}_{A}$ is only effective when adopting the mixup blending in the forgery synthesizing process.

\subsection{Adversarial Training}
%
To better leveraging the forgery augmentation space, we adopt adversarial training to dynamically construct the most challenging auxiliary task. Specifically, we use the synthesizer network $G(\cdot, \theta)$ as the generator, which maximizes the training loss of the target discriminator (\ie the detector network ${D}(\cdot, w)$) through adversarial learning. The optimization process can be presented as,
\begin{equation}
\begin{small}
\begin{aligned}
\label{all loss}
\min_{w} \max_{\theta} \mathcal{L},~~\text{s.t.}~~
\mathcal{L}(\theta,w) = \mathcal{L}_{Main} + \alpha \mathcal{L}_{R} + \mu \mathcal{L}_{T} + \gamma \mathcal{L}_{A},
\end{aligned}
\end{small}
\end{equation}
where $\alpha$, $\mu$, and $\gamma$ are weight hyper-parameters. Recall that $\theta$ denotes all the parameters in the generation process. Following the common practice of adversarial training, the above optimization problem can be approximately solved by iteratively updating the discriminator and the generator.

We first present the learning process of the discriminator. Eq.~\eqref{all loss} is a minimization problem regarding $w$. By fixing the current generator parameter $\theta$, with the learning rate of $\eta$ and batch size of $N$, we perform gradient descent update:
\begin{equation}
\begin{small}
\label{learn classfier}
w^{t+1} = w^t - \eta \frac{1}{N} \sum_{n=1}^N \nabla_{w^t} \mathcal{L}_n(\theta^t,w^t),
\end{small}
\end{equation}
where $\mathcal{L}_n$ is the loss for the $n$-th sample in a mini-batch.

The generator is designed to increase the training loss of the target discriminator by synthesizing more challenging samples than the original benign examples, thus encouraging the discriminator to learn more generalizable features. It plays a zero-sum game in an adversarial framework with the discriminator. 
Mathematically, we can formulate the object as a maximizing problem according to Eq.~\eqref{all loss},
\begin{equation}
\label{max controller}
\theta^{t+1} = arg\max_{\theta^t} \mathcal{L}(w^{t+1},\theta^t).
\end{equation}

However, directly updating $\theta$ by solving Eq.~\eqref{max controller} could be problematic because there exist non-differentiable sampling operations that will break gradient flow from ${D}(\cdot, w)$ to ${G}(\cdot, \theta)$.
To handle that, we apply REINFORCE algorithm \cite{williams1992simple} to approximate the gradient calculation for $\theta$:
\begin{equation}
\label{update A}
\begin{aligned}
\theta^{t+1} &= \theta^t + \epsilon \nabla_{\theta^t} \mathcal{L}_b\\
& \approx \theta^t + \epsilon \frac{1}{M} \sum_{m=1}^M \mathcal{L}_b \nabla_{\theta^t} \log p_m, 
\end{aligned} 
\end{equation}
where $\mathcal{L}_b = \frac{1}{N} \sum_{n=1}^N \mathcal{L}_n (w^{t+1},\theta^t)$, $M$ denotes the number of selected configuration series in a batch. $p_m$ represents the probabilities of generating $R_g$ and $T_g$, which is estimated from the sythesizer network $G(\cdot, \theta)$. 

\section{Experiments}
\subsection{Settings}
{\flushleft \bf Training datasets}.
Following recent deepfake detection methods \cite{qian2020thinking,li2020face, luo2021generalizing, liu2021spatial,wang2021representative, li2021frequency}, we train our model in the Faceforencis++ (FF++) dataset \cite{rossler2019faceforensics++}. It contains 1,000 original videos, where 720 videos are used for training, 140 videos are reserved for testing and the rest for validation.
All videos undergo four state-of-the-art deepfake methods, which includes Deepfakes (DF) \cite{df}, Face2Face (F2F) \cite{thies2016face2face}, FaceSwap (FS) \cite{fs}, and NeuralTextures (NT) \cite{thies2019deferred}. Final outputs are generated with different compression levels including RAW, High Quality (HQ) and Low Quality (LQ), respectively. 
We use the HQ and LQ versions in our experiments, and the HQ version is adopted by default unless otherwise stated.
{\flushleft \bf Testing datasets}.
To evaluate the generalizability of our method, we use the following benchmark datasets: 
CelebDF \cite{li2019celeb},  which contains 408 real videos and 795 synthesized videos that are generated by the improved deepfake technology;
Deepfake Detection Challenge (DFDC) \cite{dfdc},  which includes over 1,000 real and over 4,000 fake videos manipulated by multiple Deepfake, GANbased, and non-learned methods;
DeeperForensics-1.0 (DF1.0) \cite{jiang2020deeperforensics}, which consists of over 11,000 fake videos generated by their DF-VAE \cite{jiang2020deeperforensics} method.

We use DLIB \cite{Sagonas2016300} for face extraction and alignment, and we resize the aligned faces to $256 \times 256$ for all the samples in training and test datasets. 
{\flushleft \bf Implemenation details}.
We modify Xception \cite{chollet2017xception} as the backbones for our synthesizer and detector networks, and their parameters are initialized by Xception pre-trained on ImageNet.
The hyper-parameter used in \eqref{all loss} are $\alpha=0.1$, $\mu=0.05$, and $\gamma=0.1$.
We use the Adam optimizer \cite{kingma2014adam} for both the two networks with $\beta_1=0.9$ and $\beta_2=0.999$ and the batch size is fixed as 32. The learning rates are set to $2\times 10^{-4}$ and $5\times 10^{-5}$ for the detector and synthesizer networks, respectively.

\begin{table*}
\centering
\scalebox{0.95}{
\begin{tabular}{C{2.5cm}|C{1.1cm}C{1.1cm}C{1.1cm}|C{1.1cm}C{1.1cm}C{1.1cm}|C{1.1cm}C{1.1cm}C{1.1cm}|C{1.1cm}C{1.1cm}C{1.1cm}|C{1.1cm}}
\toprule
\multirow{2}*{Method}& \multicolumn{3}{c|}{DF} & \multicolumn{3}{c|}{F2F} & \multicolumn{3}{c|}{FS} & \multicolumn{3}{c|}{NT} & \multirow{2}*{Avg.}\\
\cline{2-13}
& \small{DFDC} & \small{CelebDF} & \small{DF1.0} & \small{DFDC} & \small{CelebDF} &\small{DF1.0} & \small{DFDC} & \small{CelebDF} &\small{DF1.0} & \small{DFDC} &\small{CelebDF} &\small{DF1.0}  \\ 
\midrule \midrule
Xception \cite{rossler2019faceforensics++} &0.654 &0.681 &0.617 &0.708 &0.598 &0.745 &0.708 &0.601 &0.605 &0.646 &0.625 &0.838 &0.669\\
Face X-ray \cite{li2020face} &0.609 &0.554 &0.668 &0.633 &0.684 &0.766 &0.646 &0.697 &\textbf{0.795} &0.613 &0.703 &0.866 &0.686\\
F3Net \cite{qian2020thinking} &0.682 &0.664 &0.658 &0.679 &0.654 &0.761 &0.679	&0.636 &0.651 &0.672 &0.689 &0.932 &0.696\\
RFM \cite{wang2021representative} &0.758 &0.723 &0.717 &0.736 &0.663 &0.732 &0.714 &0.591 &0.714 &0.726 &0.600 &0.846 &0.710\\
SRM \cite{luo2021generalizing} &0.679 &0.650 &0.720 &0.687 &0.693 &0.775 &0.671 &0.643 &0.771 &0.656 &0.651 &\textbf{0.936} &0.711\\
Ours &\textbf{0.772} &\textbf{0.730} &\textbf{0.742} &\textbf{0.787} &\textbf{0.781} &\textbf{0.786} &\textbf{0.742} &\textbf{0.800} &0.695 &\textbf{0.741} &\textbf{0.759} &0.889 &\textbf{0.769}\\
\bottomrule
\end{tabular}}
\vspace{0.1 cm}
\caption{Generalizability comparisons with state-of-the-art methods in the term of AUC. The best results are in bold. The first row denotes the training data, and the second row shows the corresponding test dataset. Our method performs favorably among the models compared.}
\label{tab result}
\vspace{-0.3 cm}
\end{table*}

\begin{table}
\centering
\scalebox{0.95}{
\begin{tabular}{c|c|C{1cm}C{1cm}|C{1cm}C{1cm}}
\hline
\multirow{3}*{Training set} & \multirow{3}*{Method} & \multicolumn{4}{c}{Test set} \\
\cline{3-6}
& & \multicolumn{2}{c|}{\small{LQ}} & \multicolumn{2}{c}{\small{HQ}}\\
\cline{3-6}
& & DF &~FS & DF &~FS \\
\midrule \midrule
\multirow{6}*{NT} &  Xception \cite{rossler2019faceforensics++} &0.587 &0.517 &0.770 &0.718\\
&Face X-ray \cite{li2020face} &0.571 &0.510 &0.585 &0.779\\
&F3Net \cite{qian2020thinking} &0.583 &0.519 &0.805 &0.612\\
&RFM \cite{wang2021representative} &0.558 &0.516 &0.798 &0.639\\
&SRM \cite{luo2021generalizing} &0.555 &0.529 &0.838 &\textbf{0.795}\\
&Ours &\textbf{0.628} &\textbf{0.568} &\textbf{0.846} &0.721\\
\bottomrule
\end{tabular}}
\vspace{0.1 cm}
\caption{Generalizability comparisons across different compression levels in the term of AUC. Our method achieves comparable performance against existing methods.}
\label{tab compression}
\vspace{-0.3 cm}
\end{table}

\subsection{Generalizability Comparisons}
To comprehensively evaluate the generalizability of our method, we compare several state-of-the-art methods including Xception \cite{rossler2019faceforensics++}, Face X-ray \cite{li2020face}, F3Net \cite{qian2020thinking}, RFM \cite{wang2021representative}, and SRM \cite{luo2021generalizing}.
To ensure fair comparisons, we use the provided codes of Xception \cite{rossler2019faceforensics++}, RFM \cite{wang2021representative}, and SRM \cite{luo2021generalizing} from the authors, and we reimplement Face X-ray \cite{li2020face} and F3Net \cite{qian2020thinking} rigorously following the companion paper's instructions and train these models under the same settings. 

{\flushleft \bf Generalizations under different datasets}.
In these experiments, we train the compared models on each of the four methods in FF++ \cite{rossler2019faceforensics++}, and evaluate it on the benchmark datasets including CelebDF \cite{li2019celeb}, DFDC \cite{dfdc}, and DF1.0 \cite{jiang2020deeperforensics}.
This setting is rather challenging because both pristines and forgeries in the test dataset are unseen in the training dataset.

We compare different methods by using the Area Under Curve (AUC) metric and present the results in Table \ref{tab result}. As seen, our method outperforms other models in most cases and achieves the overall best performance. This clearly demonstrates the advantage of the proposed adversarial augmentation and self-supervised tasks.
SRM \cite{luo2021generalizing} and F3Net \cite{qian2020thinking} both rely on high-frequency components of a image to distinguish forgeries from pristine. Our experiment suggests that they attain worse generalization performance than our approach. This may be because the high-frequency cue identified effective on the FF++ dataset \cite{rossler2019faceforensics++} may not necessarily generalize to other datasets that adopt different post-processing steps.
RFM \cite{wang2021representative} encourages the use of multiple facial regions for forgery detection and thus leads to improved generalization performance. However, their method still cannot avoid some data source biases, such as all the whole facial parts in a training sample are from the same source. This limitation may explain why their performance is inferior to ours.
Face X-ray \cite{li2020face} uses the blended artifacts in the forgery for generalization. Compared to our method, Their generalizability will degrade when these artifacts share different patterns in the training and test dataset.
As a baseline, Xception \cite{rossler2019faceforensics++} does not incorporate any augmentation or feature engineering. Its performance drops drastically in unseen forgeries. 

\begin{table}
\centering
\scalebox{1}{
\begin{tabular}{c|c|c|c}
\hline
\multirow{2}*{Training set} & \multirow{2}*{Method} & \multicolumn{2}{c}{Test set} \\
\cline{3-4}
& & F2F & FS\\
\midrule \midrule
\multirow{4}*{F2F} &  LAE \cite{du2020towards} &~~0.909~~~&~~0.632~~\\
&ClassNSeg \cite{nguyen2019multi} &0.928 &0.541 \\
&Forensic-Trans \cite{cozzolino2018forensictransfer} &0.945 &0.726 \\
&Ours &\textbf{0.960} &\textbf{0.848}\\
\bottomrule
\end{tabular}}
\vspace{0.1 cm}
\caption{Comparisons with models adopt multi-task learning in the term of ACC. Our model performs favorably against these arts.}
\label{tab mutitask}
\end{table}

\begin{table}
\centering
\scalebox{1}{
\begin{tabular}{cccc}
\hline
Method & FF++ && ~~CelebDF \\
\midrule \midrule
Two-stream \cite{zhou2017two} &0.701 &&0.538\\
Meso4 \cite{afchar2018mesonet} &0.847 &&0.548\\
MesoInception4 \cite{afchar2018mesonet} &0.830 &&0.536\\
FWA \cite{li2018exposing} &0.801 &&0.569\\
DSP-FWA \cite{li2018exposing} &0.930 &&0.646\\
Xception \cite{rossler2019faceforensics++} &0.997 &&0.653 \\
VA-MLP \cite{matern2019exploiting} &0.664 &&0.550\\
Headpose \cite{yang219exposing} &0.473 &&0.546\\
Capsule \cite{nguyen2019capsule} &0.966 &&0.575\\
SMIL \cite{li2020sharp} &0.968 &&0.563\\
Two-branch \cite{masi2020two} &0.932 &&0.734\\
SPSL \cite{liu2021spatial} &0.969 &&0.724\\
MADD \cite{zhao2021multi} &\textbf{0.998} &&0.674\\
\hline
Ours &0.984 &&\textbf{0.797}\\
\bottomrule
\end{tabular}}
\vspace{0.1 cm}
\caption{Extensive evaluations with other state-of-the-art methods in the term of AUC. The models are trained on FF++ dataset. Our method performs favorbly when tested on FF++, and it outperforms others when tested on CelebDF.}
\label{tab moreresult}
\vspace{-0.2 cm}
\end{table}

\begin{table*}[htbp]
\centering
\scalebox{0.95}{
\begin{tabular}{C{2.5cm}|C{1.1cm}C{1.1cm}C{1.1cm}|C{1.1cm}C{1.1cm}C{1.1cm}|C{1.1cm}C{1.1cm}C{1.1cm}|C{1.1cm}C{1.1cm}C{1.1cm}|C{1.1cm}}
\toprule
\multirow{2}*{Method}& \multicolumn{3}{c|}{DF} & \multicolumn{3}{c|}{F2F} & \multicolumn{3}{c|}{FS} & \multicolumn{3}{c|}{NT} & \multirow{2}*{Avg.}\\
\cline{2-13}
& \small{DFDC} & \small{CelebDF} & \small{DF1.0} & \small{DFDC} & \small{CelebDF} &\small{DF1.0} & \small{DFDC} & \small{CelebDF} &\small{DF1.0} & \small{DFDC} &\small{CelebDF} &\small{DF1.0}  \\ 
\midrule \midrule
Xception \cite{rossler2019faceforensics++} &0.654 &0.681 &0.617 &0.708 &0.598 &0.745 &0.708 &0.601 &0.605 &0.646 &0.625 &0.838 &0.669\\
Xception w/ adv &0.717 &0.703 &0.674 &0.739 &0.778 &0.735 &0.737 &0.644 &0.602 &0.662 &0.722 &0.794 &0.709\\
Ours w/ ran &0.763 &0.663 &0.690 &0.763 &0.745 &0.696 &0.738 &0.700 &0.650 &0.705 &0.666 &0.810 &0.716\\
Ours &\bf{0.772} &\bf{0.730} &\bf{0.742} &\bf{0.787} &\bf{0.781} &\bf{0.786} &\bf{0.742} &\bf{0.800} &\bf{0.695} &\bf{0.741} &\bf{0.759} &\bf{0.889} &\bf{0.769}\\
\bottomrule
\end{tabular}}
\vspace{0.1 cm}
\caption{Ablation studies regarding the effectiveness of the adversarial training. The metric is AUC. Please see Sec. \ref{sec adv} for detailed experiment settings.}
\label{tab abladv}
\vspace{-0.2 cm}
\end{table*}

\begin{table*}[htbp]
\centering
\scalebox{0.94}{
\begin{tabular}{C{2.5cm}|C{1.1cm}C{1.1cm}C{1.1cm}|C{1.1cm}C{1.1cm}C{1.1cm}|C{1.1cm}C{1.1cm}C{1.1cm}|C{1.1cm}C{1.1cm}C{1.1cm}|C{1.1cm}}
\toprule
\multirow{2}*{Method}& \multicolumn{3}{c|}{DF} & \multicolumn{3}{c|}{F2F} & \multicolumn{3}{c|}{FS} & \multicolumn{3}{c|}{NT} & \multirow{2}*{Avg.}\\
\cline{2-13}
& \small{DFDC} & \small{CelebDF} & \small{DF1.0} & \small{DFDC} & \small{CelebDF} &\small{DF1.0} & \small{DFDC} & \small{CelebDF} &\small{DF1.0} & \small{DFDC} &\small{CelebDF} &\small{DF1.0}  \\ 
\midrule \midrule
Ours w/ ran ops &0.735 &0.666 &\bf{0.782} &0.694 &0.647 &0.827 &0.737 &0.601 &0.646 &0.691 &0.685 &0.812 &0.710\\
Ours w/ ops \cite{zhang2019adversarial} &0.721 &0.726 &0.777 &0.686 &0.647 &\bf{0.840} &0.737 &0.534 &\bf{0.720} &0.678 &0.642 &0.829 &0.711\\
Ours w/ aug \cite{li2020face} &0.722	&0.692	&0.712	&0.722	&0.710	&0.739		&0.719	&0.726	&0.640 &0.714 &0.669 &0.841	&0.717 \\
Ours w/ aug \cite{zhao2021learning} &0.754 &0.679 &0.687	&0.746	&0.604	&0.753	 &0.726	&0.697 &0.674 &\bf{0.770}	&0.713	&0.863	&0.722\\
Ours &\bf{0.772} &\bf{0.730} &0.742 &\bf{0.787} &\bf{0.781} &0.786 &\bf{0.742} &\bf{0.800} &0.695 &0.741 &\bf{0.759} &\bf{0.889} &\bf{0.769}\\
\bottomrule
\end{tabular}}
\vspace{0.1 cm}
\caption{Ablation studies regarding the effectiveness of the data augmentation strategies. The metric is AUC. Please see Sec. \ref{sec aug} for detailed experiment settings.
}
\label{tab ablaug}
\vspace{-0.5 cm}
\end{table*}

\begin{table*}[htbp]
\centering
\scalebox{0.94}{
\begin{tabular}{ccc|C{1.1cm}C{1.1cm}C{1.1cm}|C{1.1cm}C{1.1cm}C{1.1cm}|C{1.1cm}C{1.1cm}C{1.1cm}|C{1.1cm}C{1.1cm}C{1.1cm}|C{1.1cm}}
\toprule
&&& \multicolumn{3}{c|}{DF} & \multicolumn{3}{c|}{F2F} & \multicolumn{3}{c|}{FS} & \multicolumn{3}{c|}{NT} & \multirow{2}*{Avg.}\\
\cline{1-15}
$\mathcal{L}_{R}$& $~~\mathcal{L}_{T}$ & $~~\mathcal{L}_{A}$ & \small{DFDC} & \small{CelebDF} & \small{DF1.0} & \small{DFDC} & \small{CelebDF} &\small{DF1.0} & \small{DFDC} & \small{CelebDF} &\small{DF1.0} & \small{DFDC} &\small{CelebDF} &\small{DF1.0}  \\
\midrule \midrule
\checkmark &\checkmark &- &0.770 &0.686 &0.687 &0.768 &0.714 &0.779 &0.722 &0.709 &0.653 &0.735 &0.720 &0.856 &0.733\\
\checkmark &- &\checkmark &0.763 &0.716 &0.709 &0.760 &0.734 &\bf{0.800} &\bf{0.776} &0.636 &\bf{0.707} &0.724 &0.683 &0.835 &0.737\\
- &\checkmark &\checkmark &0.722 &0.685 &0.656 &0.765 &0.733 &0.771 &0.713 &0.698 &0.659 &0.735 &0.709 &0.838 &0.724\\
-&-&- &0.717 &0.703 &0.674 &0.739 &0.778 &0.735 &0.737 &0.644 &0.602 &0.662 &0.722 &0.794 &0.709\\
\checkmark &\checkmark &\checkmark &\bf{0.772} &\bf{0.730} &\bf{0.742} &\bf{0.787} &\bf{0.781} &0.786 &0.742 &\bf{0.800} &0.695 &\bf{0.741} &\bf{0.759} &\bf{0.889} &\bf{0.769}\\
\bottomrule
\end{tabular}}
\vspace{0.1 cm}
\caption{Effectiveness of the proposed self-supervised auxiliary tasks. The metric is AUC. We disable the self-supervised task by assigning a zero weight to the corresponding loss.}
\label{tab ablselfsupervised}
\vspace{-0.5 cm}
\end{table*}

{\flushleft \bf Generalizations under different compression levels}.
Given that real world forgeries may be post-processed by different methods, such as compression. It is crucial that deployed deepfake detectors are not easily subverted by unseen post-processing processes.
To evaluate the generalizability of the compared models regarding the varying post-processing methods, we separately train them on NT data, while testing them on DF and FS with different compression levels.

The evaluated AUC values are presented in Table \ref{tab compression}. We can observe that almost all models generalize competitively when trained and tested on data with the same compression levels.
However, models that based on imperceptible image patterns \cite{rossler2019faceforensics++, li2020face, qian2020thinking, luo2021generalizing} suffer from large performance drop when test on unseen LQ data.
The results are not surprising since the low-level clues are largely destroyed when the images are highly compressed.
The same results are reported for RFM \cite{wang2021representative}. Because all facial parts in a face sample usually share the same compression level, exploring more facial regions cannot guarantee well generalizability across different compression levels.
On the other hand, the proposed method is substantially less affected by the compression levels, outperforming all other methods, as it uses more generalizable forgery configurations instead of just the low-level artifacts.
We believe the improvements over Xception \cite{rossler2019faceforensics++}, which is with a similar network setting, are due to the adopted adversarial self-supervised framework.

\subsection{State-of-the-art Comparisons}
{\flushleft \bf Comparison to multi-task learning detectors}.
Using multi-task learning strategies to boost deepfake detection has been explored in previous works, including LAE \cite{du2020towards}, ClassNSeg \cite{nguyen2019multi}, and Forensic-Trans \cite{cozzolino2018forensictransfer}. All these works suggest simultaneously classifying and localizing forgery regions.
Unlike their localization task, the self-supervised task involved in our self-supervised framework aims to recognize the forgery configurations selected by the synthesizer network, which not only avoids the tedious annotation work, but also considers more configurations that are prevalent in forgeries.

To ensure fair comparisons, we use the same evaluation settings in our experiments with that from the compared arts. We train our model on the F2F data and test it on F2F and FS.
The comparison results are presented in Table \ref{tab mutitask}, where our model outperforms other algorithms in test samples with both seen and unseen deepfake techniques.
Note the results are directly cited from the reported statistics in their original papers.

{\flushleft \bf Comparison to other state-of-the-art detectors}.
We further evaluate our method against several state-of-the-art models, including the Two-stream \cite{zhou2017two}, MesoNet \cite{afchar2018mesonet}, Headpose \cite{yang219exposing}, FWA \cite{li2018exposing}, VA-MLP \cite{matern2019exploiting}, Capsule \cite{nguyen2019capsule}, SMIL \cite{li2020sharp}, Two-branch \cite{masi2020two}, SPSL \cite{liu2021spatial}, MADD \cite{zhao2021multi}.
We train our model on the FF++ dataset and test it on FF++ and CelebDF. Results of some methods are directly cited from \cite{liu2021spatial}.
As shown in Table \ref{tab moreresult}, our method performs competitively with other models when tested on the FF++ dataset, and it achieves the best performance when tested on CelebDF, which further validates the effectiveness and superior generalizability of our method.

\subsection{Ablation Study}
This section analyzes the effectiveness of adversarial learning, data augmentation, and self-supervised tasks. We provide more analyses in our supplementary material.

{\flushleft \bf Adversarial Learning}.
\label{sec adv}
To validate whether the adversarial training strategy can improve the generalizability, we conduct an ablation study by comparing our methods with the following variant. (1) Xception \cite{rossler2019faceforensics++}: the baseline approach without using both adversarial learning and self-supervised learning (2) Xception w/ adv: we add adversarial augmentation (but not self-supervised learning) to the Xception baseline. (3) Ours w/ ran: we replace adversarial augmentation with random augmentation. That is, instead of relying on the forgery synthesizer to generate configurations, we randomly select configurations to perform augmentation. This variant is used to test if the improvement brought by the adversarial augmentation is from the ``adversarial training'' or the ``augmentation'' or both. 

The experimental comparison is shown in Table \ref{tab abladv}. We can see that using adversarial augmentation alone (Xception w/ adv) can lead to significant (around 4\%) improvement over the baseline approach (Xception). This shows the effectiveness of our adversarial augmentation. We also observe that if replacing the adversarial augmentation with random augmentation, the performance of our method leads to a significant drop. This suggests that adversarial training is essential for our system.

{\flushleft \bf Data augmentations}.
\label{sec aug}
%
We first replace our augmentation step with general augmentation strategies in \cite{cubuk2018autoaugment, zhang2019adversarial}, which includes 16 image operations, such as rotation and cutout \cite{devries2017improved}, and 10 magnitudes. Two augmentation forms are compared. The first one imposes random image operations on the training dataset (\ie Ours w/ ran ops); the second uses a same synthesizer network to select different operations and magnitudes for different samples, which is similar to the setting in \cite{zhang2019adversarial} (\ie Ours w/ ops \cite{zhang2019adversarial}).
Note the goals of the detector are to estimate the types of image operation and their magnitudes in these two settings.
We further compare our augmentation strategy with that from \cite{li2020face} and \cite{zhao2021learning} (\ie Ours w/ aug \cite{li2020face} and \cite{zhao2021learning}). Both these two methods suggest synthesizing new forgeries to augment the training data, where the forgery regions are fixed in the inner faces, and alpha blending is adopted for all samples in \cite{li2020face} while Poisson blending is mainly used in \cite{zhao2021learning}.

Evaluation results are shown in Table \ref{tab ablaug}.
We observe that strategies with commonly used image operations perform less effectively than our method, and the improvements gained from the adversarial training are also subtle compared to ours.
The main reason is that the commonly used image operations are used for general tasks. They may be ineffective in our task. In contrast, our augmentation method is specially designed for deepfake detection, which is more related to the forgery synthesizing process.
Thus, it is not surprising that our augmentation strategy outperforms general image operations.
Meanwhile, different from the empirically designed augmentations \cite{li2020face, zhao2021learning}, our augmentation is with multiple forms. With the help of adversarial training, more diverse samples can be created to improve the generalization. These advantages explain why our augmentation strategy performs better than \cite{li2020face, zhao2021learning}.

{\flushleft \bf Self-supervised tasks}.
We seamlessly integrate three auxiliary self-supervised tasks to boost the generalizability of our model. To evaluate the effectiveness of these tasks, we conduct ablation studies by assigning zero weight to the loss corresponds to each of these tasks (\ie $\mathcal{L}_{R}$, $\mathcal{L}_{T}$, and $\mathcal{L}_{A}$ correspond to the forgery region estimation mask, blending type estimation task, and the blending ratio estimation task, respectively).
All models are trained on the four methods of FF++ and evaluated on benchmark datasets using the same settings.
Results are shown in Table \ref{tab ablselfsupervised}. We can observe that each auxiliary task plays a vital role in our framework, where our method performs the best when all three self-supervised tasks are incorporated. On the other hand, excluding any of these tasks will decrease the overall performance, and using only the main classification loss lead to significantly worse performance than others. 
This validates the effectiveness of the proposed self-supervised tasks.

\section{Conclusions and Discussions}
In this paper, we propose a deepfake detection method that can generalize well in unseen scenarios. Our design is based on the intuition that a generalizable deepfake detector should be sensitive to different types of forgeries. We thus leverage a synthesizer and adversarial training framework to dynamically generate forgeries. Training to identify those generated forgeries, the network can learn more robust feature representation and lead to a more generalizable deepfake detector. Through extensive experiments, we demonstrate the effectiveness of the proposed method.

The major limitation of the current method is that the augmentation type in the synthesizer is still limited. One promising direction is to use GAN or other generative models to directly generate forgery images and create self-supervised auxiliary tasks. For example, the generated forgeries can be controlled by some latent variable and the auxiliary task is to predict those latent variables.

\noindent\textbf{Ethic Statement.} All face images used in this paper are adopted from existing works and are properly cited. There is no violation of personal privacy while conducting experiments in this work. 

\clearpage

{\small
\bibliographystyle{ieee_fullname}
\bibliography{detect}
}

\end{document}